\documentclass[10pt,twocolumn,letterpaper]{article}

\usepackage{iccv}
\usepackage{times}
\usepackage{epsfig}
\usepackage{graphicx}
\usepackage{amsmath}
\usepackage{amssymb}
\usepackage{multirow}
\usepackage{array}
\usepackage{booktabs}
\usepackage{bm}
\usepackage{tablefootnote}
\usepackage[para,symbol]{footmisc}
\usepackage{tikz}
\usepackage{xcolor}
\usepackage[accsupp]{axessibility}

\newcommand{\cc}{\textcolor{black}}

\usepackage[pagebackref=true,breaklinks=true,letterpaper=true,colorlinks,bookmarks=false]{hyperref}

\iccvfinalcopy 

\def\iccvPaperID{3401} 

\ificcvfinal\fi
\makeatletter
 \def\@fnsymbol#1{\ensuremath{%
    \ifcase#1
    \or 
      \dagger
    \or 
      \ddagger
    \or 
      \mathsection
    \or 
      \mathparagraph
    \else 
      \@ctrerr  
    \fi}}   
\makeatother 
\begin{document}

\title{Temporal Collection and Distribution for Referring Video Object Segmentation}

\title{Temporal Collection and Distribution for Referring Video Object Segmentation}

\author{
Jiajin Tang \quad 
Ge Zheng \quad
Sibei Yang\thanks{Sibei Yang is the corresponding author.} \vspace{2mm}\\
School of Information Science and Technology, ShanghaiTech University \\
 {\tt\small \{tangjj,zhengge,yangsb\}@shanghaitech.edu.cn} \vspace{-0mm} \\
 \small\url{https://toneyaya.github.io/tempcd}
}
\maketitle
\ificcvfinal\thispagestyle{empty}\fi

 \begin{abstract}
Referring video object segmentation aims to segment a referent throughout a video sequence according to a natural language expression. It requires aligning the natural language expression with the objects' motions and their dynamic associations at the global video level but segmenting objects at the frame level. To achieve this goal, we propose to simultaneously maintain a global referent token and a sequence of object queries, where the former is responsible for capturing video-level referent according to the language expression, while the latter serves to better locate and segment objects with each frame. Furthermore, to explicitly capture object motions and spatial-temporal cross-modal reasoning over objects, we propose a novel temporal collection-distribution mechanism for interacting between the global referent token and object queries. Specifically, the temporal collection mechanism collects global information for the referent token from object queries to the temporal motions to the language expression. In turn, the temporal distribution first distributes the referent token to the referent sequence across all frames and then performs efficient cross-frame reasoning between the referent sequence and object queries in every frame. Experimental results show that our method outperforms state-of-the-art methods on all benchmarks consistently and significantly.

 \end{abstract}

\section{Introduction}

\begin{figure}[t]
\centering
\includegraphics[width=0.99\columnwidth]{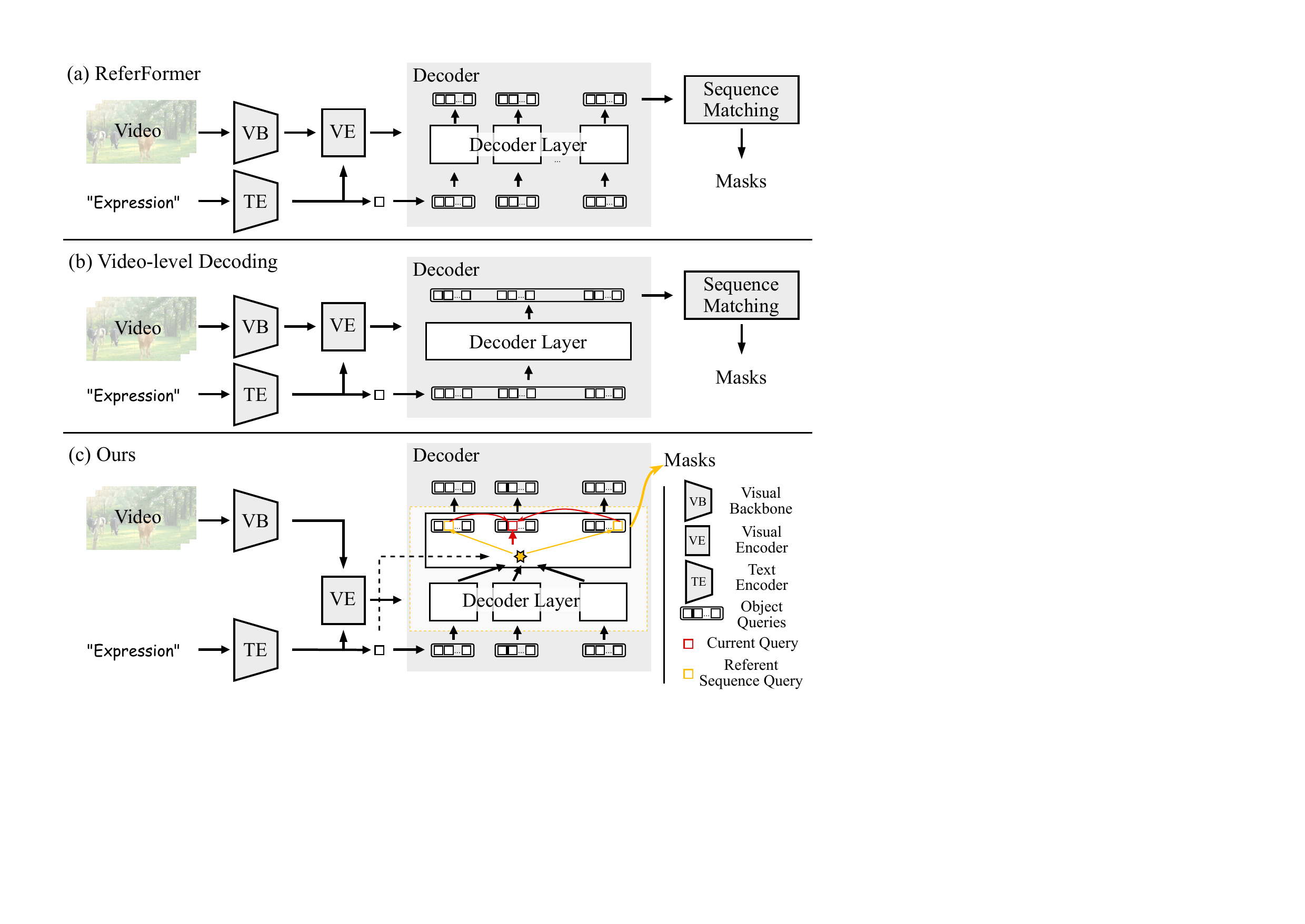}
\vspace{-0.1cm}
\caption{The comparison of (a) previous best-performing ReferFormer, (b) video-level decoding, and (c) ours referent-guided dynamic temporal decoding.}
\label{fig:intro}
\end{figure}

\begin{figure*}[t]
\centering
\includegraphics[width=1.0\textwidth]{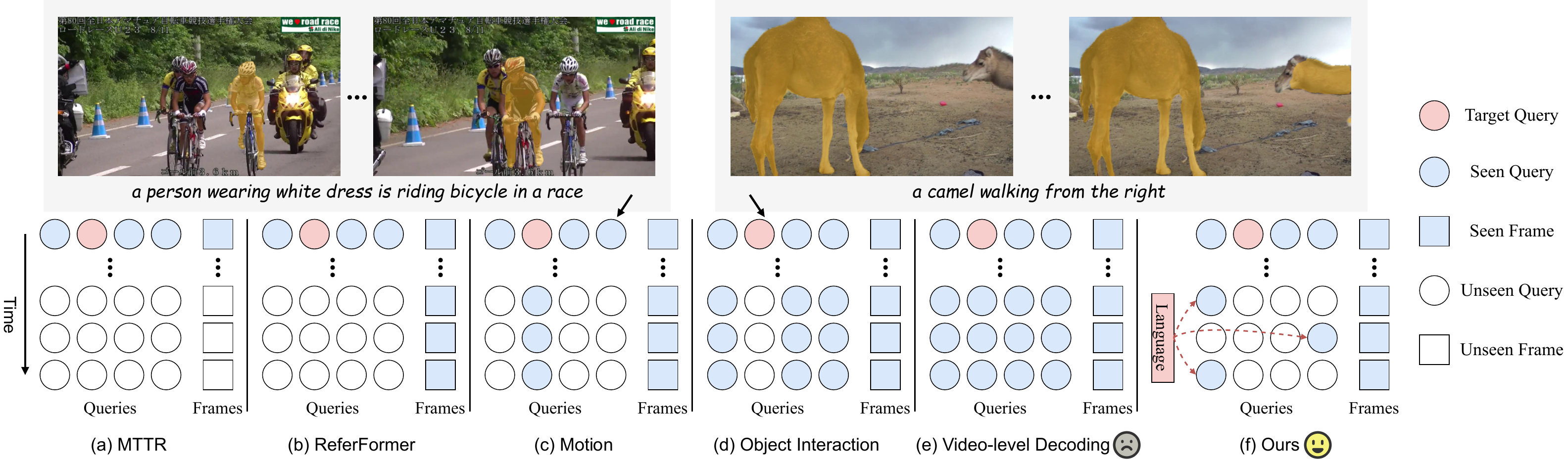}
\vspace{-0.4cm}
\caption{Comparison of temporal modeling for previous works, naive attempt, and ours. Previous works (MTTR and ReferFormer) lack temporal modeling when decoding. Besides, object motion and interaction with other objects are crucial for referring video object segmentation. However, simply modeling temporal interaction with globally visible video-level decoding cannot achieve satisfactory results. Note that our alternating interaction between the global referent and object queries can dynamically capture spatial-temporal cross-modal reasoning over objects guided by the global referent information.}
\label{fig:query}
\vspace{-0.2cm}
\end{figure*}

Referring video object segmentation task (RVOS) aims to segment the target referent throughout a video sequence given a natural language expression\cite{khoreva2019video}. 
It has attracted increasing attention from the academic community, as it is a basis for assessing a comprehensive understanding of visual, temporal and linguistic information. Meanwhile, ROVS benefits various downstream interactive applications such as language-driven human-robot interaction~\cite{qi2020reverie}, video editing~\cite{li2020manigan} and video surveillance~\cite{sreenu2019intelligent}. 
Compared to referring image segmentation~\cite{yang2021bottom, hu2016segmentation} that segments the target object in a single image mainly according to appearances and spatial and semantic relations, ROVS requires locating referents with temporal consistency according to object motions and dynamic associations among objects across frames.

Previous works mainly follow a single-stage bottom-up manner~\cite{gavrilyuk2018actor,wang2019asymmetric,wang2020context,ning2020polar,hui2021collaborative} or a two-stage top-down fashion~\cite{liang2021rethinking}. 
The bottom-up methods early align the vision and language at the local patch level, failing to explicitly model objects and their relations~\cite{liang2021rethinking, wu2022language}. Although top-down approaches explicitly extract object tracklets and select object tracklets matched with the language expression, their two-stage pipeline is complex and less efficient~\cite{wu2022language}. 
Recently, transformer-based frameworks~\cite{wu2022language, botach2022end} have been proposed to use object queries to capture objects and their associations in an end-to-end manner. 
They generate object queries and utilize the transformer decoder to search objects for these object queries in every frame, as shown in Figure~\ref{fig:intro}\textcolor{red}{a} and~\textcolor{red}{b}. 

Although these end-to-end transformer-based frameworks have achieved impressive segmentation results, they still have limitations needed to be improved. First, their object queries are frame-level and independently responsible for the search for objects in each frame, which fails to capture temporal object motions and effective object tracking, as shown in Figure~\ref{fig:query}\textcolor{red}{a} and \textcolor{red}{b}.
Therefore, we observe that they often fail to obtain temporally consistent predictions for target referents across an entire video, as shown in the left example in Figure~\ref{fig:query}.  
Second, their interactions between object queries are also solely performed per frame, leading to failure to model object-level spatial-temporal associations between objects. Therefore, they cannot correctly ground expressions that require cross-modal spatial-temporal reasoning over multiple objects. For example (see the right example in Figure~\ref{fig:query}), \textcolor{black}{they mislocate the left camel as the target object because they lack to model the relation between the left camel in the current frame and the camel walking in the subsequent frame.} 

In this paper, we aim to address these limitations by empowering the decoder with the capability of temporal modeling. 
A straightforward approach to decode objects with temporal information is to convert the frame-level decoding into video-level by inputting a sequence of object queries to search for objects in a sequence of frames like VisTR~\cite{wang2021end} for video instance segmentation, as shown in Figure~\ref{fig:intro}\textcolor{red}{b} and ~\ref{fig:query}\textcolor{red}{e}. \textcolor{black}{However, this simple attempt fails to achieve satisfactory results while significantly increasing the computational cost and even underperforming existing query-based transformer frameworks.} We further analyze the reasons: (1) The alignment between the natural language expression and the referent relies on \textit{overall objects' motions and temporal associations in the entire video}.  
For example, the two camels can be distinguished only if the walking motion of the target camel is identified based on the entire video and aligned with the language expression. 
In contrast, the above naive attempt prematurely aligns language with fine-grained frame-level objects so that its attention is distracted from focusing accurately on the overall motions and relations of objects at the video level. 
(2) However, the \textit{precise localization and segmentation of target objects should go back to and rely more on every single frame} because objects with different motions may cause them to have very different spatial locations in different frames. A similar observation is discussed by SeqFormer~\cite{wu2021seqformer} for video instance segmentation, which suggests processing attention with each frame independently. 

Therefore, to address the challenge of aligning the language expression with objects' motions and temporal associations at the global video level but segmenting objects at the local frame level, we propose to maintain both local object queries and a global referent token. Global referent token captures the video-level referent information according to the language expression, while local object queries locate and segment objects with each frame. Furthermore, the object queries and the referent token are interacted to achieve spatial-temporal object information exchange through our well-designed temporal collection and distribution mechanisms, as shown in Figure~\ref{fig:query}\textcolor{red}{f}. Specifically, the temporal collection collects the referent information from object queries with spatial and visual attributes to temporal object motions to language semantics. \textcolor{black}{In turn, the temporal distribution first dynamically distributes the referent token to every frame to extract the referent sequence. Then, the referent sequence interacts with object queries within each frame to achieve efficient spatial-temporal reasoning over objects. Note that the temporal collection and distribution alternately propagate information between object queries and the referent token to update each other.}

Finally, we propose a Temporal Collection and Distribution network (TempCD) which integrates our novel temporal collection and distribution mechanisms into the query-based transformer framework. Without using sequence matching or post-processing during inference like the previous methods~\cite{botach2022end,wu2022language}, Our TempCD can directly predict the segmentation result of every frame based on object queries in the frame and the referent token.

In summary, our main contributions are as follows,
\begin{itemize}
\setlength{\itemsep}{0pt}
\setlength{\parsep}{0pt}
\setlength{\parskip}{0pt}

    \item We introduce to maintain a global referent token and a sequence of local object queries parallelly to bridge the gap between video-level object alignment with language and frame-level object segmentation. 

    \item We propose a novel collection-distribution mechanism for interacting between the referent token and object queries to capture spatial-temporal cross-modal reasoning over objects. 
    
    \item We present an end-to-end Temporal Query Collection and Distribution (TempCD) for RVOS, which can directly predict the segmentation referent of every frame without any post-processing and sequence matching. 

    \item Experiments on Ref-Youtube-VOS, Ref-DAVIS17, A2D-Sentences and JHMDB-Sentences datasets show that TempCD outperforms state-of-the-art methods on all benchmarks consistently and significantly.
    
\end{itemize}

\section{Related Work}
\noindent \textbf{Referring Image Segmentation} (RIS) involves segmenting objects in images based on natural language expressions. Compared to Referring Video Object Segmentation (RVOS), RIS operates on individual images without temporal information. Previous works~\cite{li2018referring,chen2019referring,chen2019see,cgformer,lin2021structured,yang2020propagating,yang2019cross,yang2019dynamic,chengshi} focus on the joint modeling of vision and language. Various methods~\cite{ding2021vision,jiao2021two,jing2021locate,ye2020dual,huang2020referring,hu2020bi,luo2020multi,hui2020linguistic,yang2020relationship} are explored successively, such as using fusion operators like concatenation, ConvLSTM, attention mechanisms~\cite{vaswani2017attention}, and GNN~\cite{gori2005new} to obtain multimodal semantic feature maps. In addition, some studies attempt to decouple different components or key semantics from language, through explicit two-stage approaches~\cite{yu2018mattnet,wu2020phrasecut} or implicit attention modules~\cite{shi2018key,ye2019cross,ye2020dual,yang2021bottom}, to achieve a more fine-grained understanding. Recent models explore novel fusion techniques, including early fusion~\cite{yang2022lavt,feng2021encoder}, Linguistic Seed Encoder~\cite{kim2022restr}, and contrastive learning~\cite{wang2022cris}, to improve the performance of RIS. 
However, for RVOS, it requires not only the multimodal integration of language and image frames but also temporal modeling at the video level. 

\begin{figure*}[t]
\centering
\includegraphics[width=0.9\textwidth]{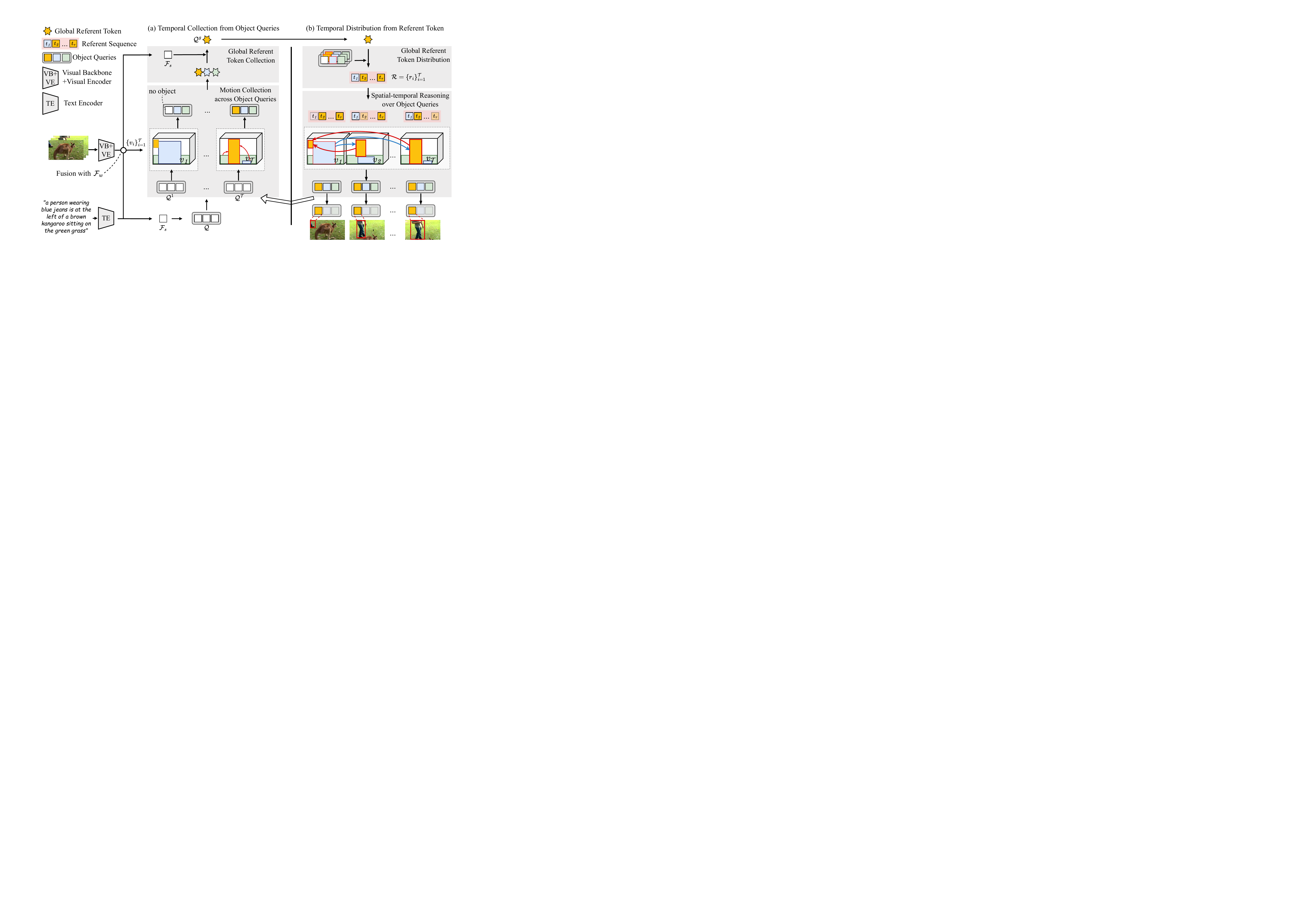}
\caption{Overall framework of \cc{TempCD}. \cc{TempCD} simultaneously maintains a global referent token and a sequence of object queries, and they interact via the proposed collection-distribution mechanism. Specifically, the temporal collection collects global information for the referent token from object queries to the temporal motions to the language expression. In turn, the temporal distribution first distributes the referent token to the referent sequence and then performs efficient cross-frame reasoning between the referent sequence and object queries.}
\vspace{-0.2cm}
\label{fig:framework}
\end{figure*}
\noindent \textbf{Referring Video Object Segmentation} (RVOS) aims to segment the target object described in natural language from videos. Previous works in RVOS has primarily used two frameworks: bottom-up and top-down. The top-down framework~\cite{liang2021rethinking} directly models the motion information between segmentation masks of consecutive frames. However, this two-stage approach, which includes complex computational costs, is limited in terms of video input length. For the bottom-up framework, some approaches~\cite{gavrilyuk2018actor,wang2019asymmetric,wang2020context,ning2020polar,hui2021collaborative} directly apply RIS methods to construct multi-modal feature maps for referring segmentation of keyframes. They rely solely on replacing the traditional image backbone with a 3D temporal backbone for temporal modeling, which limits the performance of multi-frames segmentation. To address this issue, URVOS~\cite{seo2020urvos} extends previous methods by building a memory bank that propagates language referent object information in the temporal dimension. Furthermore, URVOS introduces the Ref-Youtube-VOS~\cite{seo2020urvos} dataset, which provides segmentation annotations for each frame that require more efficient temporal modeling. In addition, recent works~\cite{wu2022multi,ding2022language, zhao2022m4} adopt language guided fusion between temporal features and visual features to obtain more efficient temporal modeling. Recently, query-based methods~\cite{wu2022language,botach2022end} with a transformer encoder-decoder frameworks~\cite{cheng2021per,carion2020end,zhu2020deformable} attempt to capture object-level information. They achieve higher performance by constructing a multimodal feature map of visual and linguistic information, or by using language-initialized queries with independently decoding with queries for each frame.

\noindent \textbf{Temporal Modeling in Video Instance Segmentation} (VIS). \textcolor{black}{VIS aims to simultaneously track, segment and classify interest instances in a video.} Similarly to RVOS, VIS requires obtaining instance information from every frame in the video and maintaining the temporal consistency of instances. Recent works focus on spatial and temporal modeling based on instance-level information. \textcolor{black}{VisTR~\cite{wang2021end} employs an encoder-decoder transformer based on DETR, with a concateneated sequence of instance queries as the input to enable spatial-temporal interactions.} \textcolor{black}{The following IFC~\cite{hwang2021video} and TeViT~\cite{yang2022temporally} explored a more efficient approach to temporal modeling in the encoder by introducing supplementary tokens for interaction across frames, with the decoder aligned with VisTR. } \textcolor{black}{Other following works~\cite{xu2021instance, koner2022instanceformer} take advantage of memories or queries from several previous frames to help instance segmentation in the current frame, extending to a online fashion.} \textcolor{black}{Recently, SeqFormer~\cite{wu2021seqformer} proposes that the acquisition of frame-level instance information such as position needs to go back to each frame, i.e., the decoding process should be done independently for each frame.  While these methods provide potential solutions for temporal modeling, the direct application to RVOS is not feasible, because of inconsistent object semantics over frames.
}


\section{Method}
The framework of our proposed \cc{TempCD} is shown in Figure~\ref{fig:framework}. 
We introduce a sequence of object queries to capture object information in every frame independently and maintain a referent token to capture the global referent information for aligning with language expression. The referent token and object queries are alternately updated to stepwise identify the target object and better segment the target object in every frame through the proposed temporal collection and distribution mechanisms.

Specifically, we first introduce encoders and \cc{the definition of  object queries} in Section~\ref{sec:3.1}. Then, we present the temporal collection that collects the referent information for the referent token from object queries to temporal object motions to the language expression (see Section~\ref{sec:3.2}). Next, we introduce the temporal distribution mechanism that distributes the global referent information to the referent sequence across frames to object queries in each frame (see Section~\ref{sec:3.3}). The collection-distribution mechanism explicitly captures object motions and spatial-temporal cross-modal reasoning over objects. Finally, we introduce segmentation heads and loss functions in Section~\ref{sec:3.4}. Note that as we explicitly maintain the referent token, we can directly identify the referent object in each frame without requiring sequence matching as in previous queries-based methods.

\subsection{Encoders and Query Definition}
\label{sec:3.1}
\noindent \textbf{Language Encoder.}
Following previous work~\cite{wu2022language}, we adopt RoBERTa~\cite{liu2019roberta} as our language encoder. Given a language expression with $L$ words, we extract its sentence feature $\mathcal{F}_s \in \mathbb{R}^C$, where $C$ denotes channel dimensions. In addition, we obtain the output feature representation of each word before the last pooling layer as $\mathcal{F}_{w} \in \mathbb{R}^{L \times C}$.  

\noindent \textbf{Visual Backbone and Encoder.}
Following previous works~\cite{wu2022language, wu2022multi, botach2022end, yang2022temporally}, we use ResNet-50~\cite{he2016deep} as our visual backbone. We also follow recent work~\cite{wu2022language, botach2022end} to evaluate our model using the temporal visual backbone network Video-Swin~\cite{liu2021video} in experiments. For an input video $V \in \mathbb{R}^{H\times W\times 3\times \mathcal{T}}$ with $\mathcal{T}$ frames, we first extract its visual feature maps and further construct multimodal feature maps $\mathcal{V}=\{v_i\}_{i=1}^\mathcal{T}$ by fusing the visual feature maps with the word features of the expression $\mathcal{F}_{w}$ following~\cite{wu2022language}.

\noindent \textbf{Object Query Definition.}
Inspired by query-based detection and segmentation frameworks~\cite{carion2020end, zhu2020deformable, cheng2021per}, we leverage object queries to represent object-level information. For the video with $\mathcal{T}$ frames, we introduce a sequence of $\mathcal{T} \times N$ object queries to represent the objects in every frame and follow~\cite{wu2022language} to use the language feature $\mathcal{F}_s$ to initialize the object queries as $\mathcal{Q} \in\mathbb{R}^{\mathcal{T}\times N\times C}$, where the number of objects and the dimension of channels are $N$ and $C$, respectively. 

\noindent \textbf{Referent Token Definition.} In addition to adopting the object queries to represent objects in every frame, we define a referent token $\mathcal{Q}^g\in\mathbb{R}^C$ to represent the global information of the target object. As the language expression naturally indicates the referent information, we thus use the language feature $\mathcal{F}_s$ to initialize the $\mathcal{Q}^g$.

\subsection{Temporal Collection from Object Queries}
\label{sec:3.2}
The temporal collection mechanism collects object motions from frames and aggregates object motions consistent with the language expression into the global referent token. 
We first perform motion collection across object queries (see Section~\ref{sec:3.2.1}) and then update the global referent token from the object motions and language expression (see Section~\ref{sec:3.2.2}). 

\subsubsection{Motion Collection across Object Queries}
\label{sec:3.2.1}
For simplicity of demonstration, we denote the object queries output from the $(l-1)$-th layer as $\mathcal{Q}_{l-1}$ and input them into the $l$-th layer of \cc{TempCD}. Note that we input the object queries $\mathcal{Q}$ defined in Section~\ref{sec:3.1} into the first layer of \cc{TempCD}. 
We utilize $\mathcal{Q}_{l-1}$ to capture object information in each frame and integrate them to obtain temporal information such as object motion and action. 
Specifically, for $i$-th frame, we feed its object queries $q_i=\mathcal{Q}_{l-1}^i\in\mathbb{R}^{N\times C}$ and the multimodal features $v_i$ into the DETR transformer decoder~\cite{carion2020end, zhu2020deformable} to locate objects, where $\mathcal{Q}_{l-1}^i$ means the object queries of $i$-th frame.  The computation proceeds as follows,
\begin{equation}
\begin{aligned}
    \hat{\mathcal{Q}}_{l} &= \{\mathrm{DETR}(q_i,v_i)\}_{i=1}^\mathcal{T},
\end{aligned}
\end{equation}
\noindent where $\mathrm{DETR }(\cdot,\cdot)$ represents the DETR transformer decoder, and $\hat{\mathcal{Q}}_{l}$ denotes the object queries output by the decoder. 

Next, to aggregate object motions across the video sequence, we combine objects information from queries of each frame. Specifically, we generate temporal weights $G \in\mathbb{R}^{\mathcal{T}\times N \times 1}$ for each frame given the object queries $\hat{\mathcal{Q}}_{l}$. This is done by employing a nonlinear layer followed by a softmax operator. Then, we use the temporal weights $G$ to integrate the frame-level information and obtain object motions $M\in\mathbb{R}^{N\times C}$ through a gating mechanism. The integration is computed as follows,
\begin{equation}
\label{eq:gate}
\begin{aligned}
	&G = \mathrm{softmax}(\hat{\mathcal{Q}}_lW),\\
	&M = \sum^{\mathcal{T}}_{i=1}(G_i\odot \hat{\mathcal{Q}}_l^i),
\end{aligned}
\end{equation}
where $W\in \mathbb{R}^{C\times1}$ denotes the learnable weights of the linear layer, $G_i \in \mathbb{R}^{N\times 1}$ is the weight for $i$-th frames in $G$, $\mathrm{softmax}(\cdot)$ represents the softmax operation along the temporal dimension, and $\odot$ is the element-wise product. 

\subsubsection{Global Referent Token Collection}
\label{sec:3.2.2}
Next, we update the global referent token $\mathcal{Q}_{l-1}^g \in \mathbb{R}^C$ at the ($l-1$)-th layer to referent token $\mathcal{Q}_{l}^g \in \mathbb{R}^C$ by selecting the object motions $M\in\mathbb{R}^{N\times C}$ consistent with the language feature $\mathcal{F}_s \in \mathbb{R}^C$. Note that we input the global referent token $\mathcal{Q}^g$ defined in Section~\ref{sec:3.1} into the first layer of ~\cc{TempCD}.
We first calculate the cosine similarity between the language feature $\mathcal{F}_s$ and the object motions $M$. We obtain the similarity scores $S=\{s_j\}_{j=1}^{N}$ as follow,
\begin{equation}
\label{eq:cos}
\begin{aligned}
	s_j &= \mathrm{cos}(\mathcal{F}_s, m_{j}),\\
\end{aligned}
\end{equation}
where $\mathrm{cos(\cdot,\cdot)}$ means the cosine similarity function and $m_{j}$ denotes the $j$-th object's motion feature in $M$. Next, we update the global referent token $\mathcal{Q}_g$ as the motion feature of the object with the highest similarity score to the global referent token. To ensure the selection process is differentiable, we implement this selection through the Gumbel Softmax operator~\cite{jang2016categorical,maddison2016concrete}. The calculation is as follows:
\begin{equation}
\begin{aligned}
	&S_{\text{gumbel}}=\mathrm{softmax}((S+G_s)/\tau),\\
	&S_{\text{onehot}}=\mathrm{onehot}(\mathrm{argmax}_N(S_{\text{gumbel}})),\\
	&S_{\text{referent}}=S_{\text{onehot}}-\mathrm{sg}[S_{\text{gumbel}}]+S_{\text{gumbel}},
\end{aligned}
\end{equation}
where $\mathrm{softmax}(\cdot)$ represents the softmax operation along dimension $N$, $G_s\in\mathbb{R}^{N\times1}$ is sampled from the $\mathrm{Gumbel}(0,1)$ distribution, $\tau$ is the temperature coefficient hyperparameter, and $\mathrm{sg}$ is the stop gradient operation. Additionally, $\mathrm{argmax}_N$ represents choosing the index with the highest similarity score, and $\mathrm{onehot}$ represents setting the highest-scoring item to 1 based on the selected index and setting the others to 0. The one-hot vector $S_{\text{referent}} \in \mathbb{R}^{N\times 1}$ indicates the selection from the object motions. The global referent token $\mathcal{Q}^g \in \mathbb{R}^{C}$ is updated as follows:
\begin{equation}
	\mathcal{Q}^g_{l} = \mathrm{MLP}(\mathrm{LayerNorm}(S_{\text{referent}}^{\top}M)+Q^g_{l-1}),
\end{equation}
where $\mathrm{MLP}(\cdot)$ denotes a three-layer linear layer with activation function, and $\mathrm{LayerNorm(\cdot)}$ is the same as the LayerNorm in standard Transformer. 

We obtain the updated global referent token $\mathcal{Q}^g$ that contains information on the referent object aligned with the language expression. Next, we feed it into the \cc{Temporal Distribution} module to assist in distributing the referent information into each frame.

\subsection{Temporal Distribution from Referent Token}
\label{sec:3.3}
The temporal distribution aims to propagate the referent information to all frames and perform dynamically spatial-temporal reasoning over objects based on the referent information. First, we distribute the global referent token to each frame independently to extract the referent sequence across all frames (see Section~\ref{sec:3.3.1}). Next, we perform the efficient cross-frame reasoning between the referent sequence and object queries in every frame (see Section~\ref{sec:3.3.2}).

\subsubsection{Global Referent Token Distribution}
\label{sec:3.3.1}
Given the global referent token $\mathcal{Q}_l^g$, we distribute the referent object information to every frame and further obtain the referent sequence across all frames, denoted by $\mathcal{R}_l=\{r_i\}_{i=1}^\mathcal{T}$, which refers to the target object in each frame.

Specifically, we first compute the cosine similarity scores between the object queries $\hat{\mathcal{Q}}_{l} = \{q_i\}_{i=1}^\mathcal{T}$ and the global referent token $\mathcal{Q}^g_{l}$. Then we obtain the index $d_i$ of the object query with the highest similarity score to the global referent token for every frame $i$ independently. The computation is as follows,

\begin{equation}
\begin{aligned}
	&s_{i,j} = \mathrm{cos}(q_{i,j},\mathcal{Q}^g_{l}),\\
	&d_i=\mathrm{argmax}_{j}\{s_{i,j}\},
\end{aligned}
\end{equation}
where $q_{i,j}$ is the $j$-th object query in the $i$-th frame, and $s_{i,j}$ is the corresponding similarity score. For every frame $i$, we select the object query $d_i$ with the highest similarity score and fuse it with the global referent token $\mathcal{Q}^g_{l}$ to generate the referent sequence $\mathcal{R}_l=\{r_i\}_{i=1}^\mathcal{T}$ as follows,
\begin{equation}
	\mathcal{R}_l=\{\mathrm{MLP}(\mathrm{LayerNorm}(q_{i,d_i}+\mathcal{Q}^g_{l}))\}_{i=1}^ \mathcal{T}.
\end{equation}
We end up with referent sequence $\mathcal{R}_l\in\mathbb{R}^{T\times C}$ across all frames, which is used as crucial objects to interact with object queries in every frame independently. 

\subsubsection{Spatial-temporal Reasoning over Object Queries}
\label{sec:3.3.2}

We leverage the referent sequence $\mathcal{R}_l\in\mathbb{R}^{T\times C}$, which encodes the referent object information across all frames, to interact with the object queries to achieve the cross-frame temporal reasoning over multiple objects and further distribute referent information to all object queries in all frames. 

Specifically, for each object query $q_{i,j}$ at the $i$-th frame, we concatenate it with the objects of the referent sequence in all other frames to construct the cross-object query sequence $C_{i,j}=\{r_1,..,r_{i-1},q_{i,j},r_{i+1}...,r_{\mathcal{T}}\} \in \mathbb{R}^{\mathcal{T}\times C}$. The object-level interaction across frames is implemented by multi-head self attention, which is computed as follows,

\begin{equation}
\begin{aligned}
	\hat{C_{i,j}}&=\mathrm{MHSA}(C_{i,j}), \\
\end{aligned}
\end{equation}
where $\mathrm{MHSA}(\cdot)$ denotes the multi-head self-attention mechanism. 
The $j$-th updated object queries in $\hat{C_{i,j}}$ for every frame $i$ are concatenated as $\mathcal{Q}^f_{l}\in\mathbb{R}^{{\mathcal{T}\times N\times C}}$ to input to next decoder layer. 
Note that our cross-frame interaction only requires interaction with referent sequences to object queries in each frame, which achieves efficient cross-frame interaction over multiple objects.

\subsection{Prediction Heads and Loss}
\label{sec:3.4}
We integrate the referent sequence and object queries in each decoder layer to the prediction heads for predicting results. To simplify the presentation, we use the final layer for details.

We extract the referent sequence $\mathcal{R}\in\mathbb{R}^{T\times C}$ and combine them with the object queries $Q_l$ to obtain the output referent sequence $\hat{\mathcal{R}}$ for prediction. Specifically, we use a gating mechanism to integrate the information from $\hat{C}$ into video-level features, which is similar to Eq~\ref{eq:gate}. Next, we calculate cosine similarity between each sequence $C_{j,i}$ and the language sentence feature $\mathcal{F}_s$, which is similar to Eq~\ref{eq:cos}, to obtain the score for each object $q_{i,j}$. We select the highest-scoring object ${q^r_i}$ in each frame and combine it with $\mathcal{R}$ to obtain the final referent sequence $\hat{\mathcal{R}} \in \mathbb{R}^{\mathcal{T}\times C}$ of the referred object in each frame:
\begin{equation}
	\hat{\mathcal{R}} = \{\mathrm{MLP}(\mathrm{LayerNorm}(r_i+q^r_i))\}.
\end{equation}
Finally, we feed $\hat{\mathcal{R}}$ into the segmentation head to predict the segmentation results. Additionally, similar to ~\cite{wu2022language} and ~\cite{botach2022end}, we use a box prediction head to regress the box coordinates for each query. \cc{We utilize Dice~\cite{milletari2016v} loss and Focal~\cite{lin2017focal} loss as the segmentation loss and GIOU~\cite{rezatofighi2019generalized} loss and L1 loss as the box loss.}
\begin{table*}[t]
\small
\begin{center}
\renewcommand\arraystretch{0.88}
\setlength{\tabcolsep}{3.6mm}{\begin{tabular}{c|c|ccc|ccc}
\toprule[1pt]
\rule{0pt}{10pt} 
\multirow{2}{*}{Backbone}  &
\multirow{2}{*}{Method} & 
\multicolumn{3}{c|}{Ref-Youtube-VOS}  &
\multicolumn{3}{c}{Ref-Davis17} \\
\cline{3-8}
\rule{0pt}{10pt} 
 && $\mathcal{J}$& $\mathcal{F}$& \bm{$\mathcal{J}\&\mathcal{F}$}& $\mathcal{J}$& $\mathcal{F}$& \bm{$\mathcal{J}\&\mathcal{F}$} \\
\hline
\noalign{\smallskip}
\multirow{7}{*}{ResNet-50}
 &CMSA~\cite{ye2019cross}  &33.3 &36.5&34.9  &32.2 &37.2 &34.7  \\
 &CMSA+RNN~\cite{ye2019cross}   &34.8 &38.1&36.4  &36.9 &43.5&40.2  \\
 &URVOS~\cite{seo2020urvos}   &45.3 &49.2&47.2  &47.3 &56.0&51.5  \\
&MLRL~\cite{wu2022multi} &48.4 &51.0 &49.7 &50.1 &55.4 &52.7 \\
  &LBDT~\cite{ding2022language}  &48.2 &50.6 &49.4 &-- &-- &54.5 \\
    &ReferFormer~\cite{wu2022language} &54.8 &56.5 &55.6 &55.8 &61.3 &58.5 \\
 &\textbf{Ours}   &\textbf{57.5}   &\textbf{60.5}  &\textbf{59.0} &\textbf{57.3} &\textbf{62.7}   &\textbf{60.0} \\
\hline
\noalign{\smallskip}
\multirow{3}{*}{Video-Swin-T}
 &MTTR~\cite{botach2022end}  &54.0   &56.6  &55.3 &-- &-- &--  \\ 
 &ReferFormer~\cite{wu2022language}  &58.0   &60.9  &59.4 &-- &-- &--  \\
 &\textbf{Ours}   &\textbf{60.5}   &\textbf{64.0}  &\textbf{62.3} &\textbf{59.3} &\textbf{65.0}   &\textbf{62.2}    \\  
\hline
\noalign{\smallskip}
\multirow{2}{*}{Video-Swin-B}
 &ReferFormer~\cite{wu2022language}  &61.3   &64.6  &62.9 &58.1 &64.1 &61.1  \\
 &\textbf{Ours}   &\textbf{63.6}   &\textbf{68.0}  &\textbf{65.8} &\textbf{61.6} &\textbf{67.6}   &\textbf{64.6}    \\  
\bottomrule[1pt]
\end{tabular}}
\end{center}
\caption{Comparison with state-of-the-art models for RVOS on Ref-Youtube-VOS and Ref-Davis-2017 datasets.}
\label{tab:youtube}
\vspace{-0.2cm}
\end{table*}

\begin{table}[t]
\begin{center}
\small
\renewcommand\arraystretch{0.95}
\setlength{\tabcolsep}{1.0mm}{
\begin{tabular}{c|c|cc|cc}
\toprule[1pt]
\multirow{2}{*}{Backbone}  & \multirow{2}{*}{Method} & \multicolumn{2}{c|}{A2D-S} & \multicolumn{2}{c}{JHMDB-S}  \\
\cline{3-6}
\rule{0pt}{6pt} 
 && oIoU& mIoU & oIoU& mIoU  \\
\hline
\noalign{\smallskip}
\multirow{1}{*}{VGG16}
&Hu \etal~\cite{hu2016segmentation} &47.4&35.0&54.6&52.8 \\
\noalign{\smallskip}
\hline
\noalign{\smallskip}
\multirow{3}{*}{I3D}
&Gavrilyuk \etal~\cite{gavrilyuk2018actor} &53.6&42.1&54.1&54.2  \\
&ACAN \cite{wang2019asymmetric}  &60.1&49.0&75.6&56.4 \\
&CMPC-V \cite{liu2021cross}  &65.3&57.3&61.6&61.7 \\
\noalign{\smallskip}
\hline
\noalign{\smallskip}
\multirow{4}{*}{Resnet-50}
    &ClawCraneNet \cite{liang2021clawcranenet}  &63.1&59.9&64.4&65.6 \\
    &MMMMTBVS~\cite{zhao2022m4} &67.3&55.8&61.9&61.3 \\
    &LBDT \cite{ding2022language}  &70.4&62.1&64.5&65.8 \\
    &\textbf{Ours}   &\textbf{76.6}   &\textbf{68.6}&\textbf{70.6}   &\textbf{69.6}  \\  
\bottomrule[1pt]
\end{tabular}}
\end{center}
\caption{Comparison with state-of-the-art models for RVOS on A2D-Sentences (A2D-S) and JHMDB-Sentences (JHMDB-S).}
\label{tab:a2d}
\vspace{-0.4cm}
\end{table}
\section{Experiment}
\noindent\textbf{Datasets.} We conduct experiments on four benchmark datasets that are publicly available: Ref-Youtube-VOS~\cite{seo2020urvos}, Ref-Davis-2017~\cite{khoreva2019video}, A2D-Sentences~\cite{gavrilyuk2018actor}, and JHMDB-Sentences~\cite{gavrilyuk2018actor}. Ref-Youtube-VOS consists of 3978 videos and approximately 15K language descriptions. Ref-Davis-2017 contains approximately 90 videos. A2D-Sentences and JHMDB-Sentences~\cite{xu2015can, jhuang2013towards}, originally focused on action recognition, have been expanded to incorporate language expression annotations. This expansion has yielded a total of 3782 videos for A2D-Sentences and 928 videos for JHMDB-Sentences, each accompanied by corresponding language descriptions. 

\noindent\textbf{Implementation Details.} Following~\cite{wu2022language}, we pre-train our model on RefCOCO dataset~\cite{yu2016modeling}. We utilize a text encoder derived from RoBERTa~\cite{liu2019roberta}, coupled with either ResNet50~\cite{he2016deep} or Video-Swin serving~\cite{liu2021video} as our visual backbone. We train our model for 6 epochs with an initial learning rate of 1e-4 and the AdamW~\cite{loshchilov2017adamw} optimizer. Consistent with prior works~\cite{seo2020urvos, wu2022language, botach2022end}, we use the $\mathcal{J}$, $\mathcal{F}$, and $\mathcal{J}\&\mathcal{F}$ metrics for evaluation on Ref-Youtube-VOS and Ref-Davis-2017 datasets, and Overall IoU and Mean IoU as evaluation metrics on A2D-Sentences and JHMDB-Sentences datasets.

\subsection{ Comparison with State-of-the-Art Methods}

As shown in Table~\ref{tab:youtube} and Table~\ref{tab:a2d}, we compare TempCD with state-of-the-art methods on four benchmarks. TempCD consistently outperforms state-of-the-art methods on all datasets. 

\noindent\textbf{Comparison of Ref-Youtube-VOS and Ref-Davis-2017 Datasets.} \cc{Results for the Ref-Youtube-VOS and Ref-Davis-2017 datasets are presented in Table~\ref{tab:youtube}. With a standard ResNet-50~\cite{he2016deep} visual backbone, our model exhibits improvements of 2.7\%, 4\%, and 3.4\% for the $\mathcal{J}$, $\mathcal{F}$, and $\mathcal{J}\&\mathcal{F}$ metrics, respectively, on the Ref-YoutubeVOS dataset. Employing the advanced temporal visual backbone, Video-Swin-B~\cite{liu2021video}, our approach consistently outperforms the previous state-of-the-art model~\cite{wu2022language} on the Ref-Davis-2017 dataset, achieving a 3.5\% increase across the aforementioned metrics. Note that evaluations on the RefDavis-2017 dataset are performed using models trained on the Ref-Youtube-VOS dataset.}

\cc{Furthermore, we present a thorough evaluation of the effectiveness of our TempCD through subsequent comparisons: (1) We compare our model to MTTR~\cite{botach2022end}, a pioneering method that adopts a query-based framework for RVOS, despite lacking temporal modeling in both the encoder and decoder stages. Our TempCD surpasses MTTR, achieving a 7\% improvement in the $\mathcal{J\&F}$ metric on the Ref-Youtube-VOS dataset. This outcome underscores the heightened efficacy of our introduced temporal decoding modules.
(2) Compared with LBDT~\cite{ding2022language}, our TempCD achieves a significant 9.6\% enhancement in the $\mathcal{J\&F}$ metric. This result suggests that, unlike \cite{ding2022language, wu2022multi, seo2020urvos} that integrate temporal and visual features via a bottom-up framework, our proposed TempCD achieves more efficient temporal modeling.}

\begin{figure*}[t]
\centering
\includegraphics[width=0.99\textwidth]{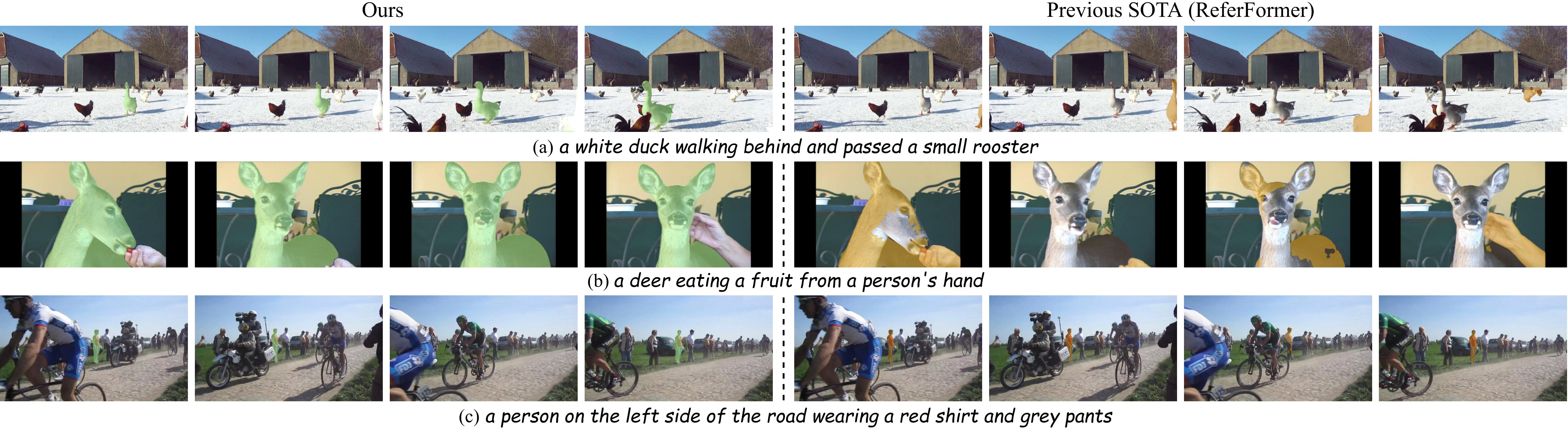}
\caption{Visualization for segmentation results of previous best-performing ReferFormer~\cite{wu2022language} and ours.}
\label{fig:vz}
\vspace{-0.2cm}
\end{figure*}

\noindent\textbf{Comparison on A2D-Sentences and JHMDB-Sentences.} 
\cc{As presented in Table~\ref{tab:a2d}, our approach yields mean improvements of 6.2\% in Overall IoU and 5.2\% in Mean IoU, surpassing the predominant state-of-the-art method~\cite{ding2022language} for these two datasets. In contrast to Ref-Youtube-VOS, these datasets predominantly comprise segmentation annotations for keyframes that encapsulate actions. When compared to prior studies~\cite{gavrilyuk2018actor,wang2019asymmetric,wang2020context,ning2020polar,hui2021collaborative} that adopt temporal encoders to process video temporal information, our method consistently attains enhancements across these two datasets. This highlights the significance of explicitly capturing and integrating video-level temporal context into individual frames.}

\subsection{Ablation Study}
To validate the impact of modules in our model, we evaluate six variants of our TempCD on Ref-Youtube-VOS dataset. The results are shown in Table~\ref{tab:ab}. All experiments are performed with Video-Swin-B as the visual backbone. 

\noindent \textbf{Baselines.}
\cc{(1) Our baseline model employs $\mathcal{T}$ sets of frame-level object queries (``Local Query"). These queries are concatenated sequentially and fed into the decoder, following a similar approach to VisTR~\cite{wang2021end}. The derived results (53.7\% $\mathcal{J}$, 56.3\% $\mathcal{F}$, and 55.0\% $\mathcal{J}\&\mathcal{F}$) suggest a misalignment between the natural language representations and the referent semantics at the individual frame level. (2) Another subsequent baseline is established by integrating a singular shared set of video-level object queries across all frames (``Global Query"). The global queries are performed an inner product with the feature map $v$ of each frame, yielding segmentation outcomes. While this approach intrinsically aligns with language representations and achieves consistency between frames, it struggles to adapt to frame-level variability, resulting in a performance decline relative to the first baseline.}

\noindent \textbf{Query Collection and Distribution.}
(3) A natural enhancement for baseline (1) entails integrating cross-frame temporal attention on a referent
sequence, which is selected from local queries by the similarity with language. This modification leads to a performance improvement of 3.4\%.
(4) Another endeavor to enhance the baseline (1) involves the introduction of an additional set of global queries. These queries collect motion information from local queries and serve to distribute global context to local queries. Specifically, we utilize a Top-K strategy to select global information based on the similarity score with language. This facilitates implicit frame level interaction and, significantly, fosters semantic alignment between global queries and language expressions. This alignment contributes to a marked performance gain of 5.6\%.
(5) We further incorporate both (3)'s explicit cross-frame temporal modeling and (4)'s motion collection and distribution, which delivers a 2.8\% improvement.
(6) We replace the Top-K selection, which does not allow gradient calculation, with Gumbel Softmax, improving $\mathcal{J}\&\mathcal{F}$ by 0.8\%. 
(7) Our complete model further enhances the vanilla cross-frame temporal attention mechanism. Instead of solely updating the referent sequence via temporal interaction, local object queries are also extended to incorporate temporal context information from referent sequence, facilitating cross-frame interaction over multiple objects. The full decoder of our proposed method achieves the performance of 65.8\% in terms of $\mathcal{J}\&\mathcal{F}$.

\begin{table}[t]
\small
\begin{center}
\renewcommand\arraystretch{0.99}
\setlength{\tabcolsep}{2.5mm}{
\begin{tabular}{c|cccc}
\toprule[1pt]
\rule{0pt}{6pt} 
&Method& \phantom{10}$\mathcal{J}$\phantom{10}& \phantom{10}$\mathcal{F}$\phantom{10}& \bm{$\mathcal{J}\&\mathcal{F}$}\\
\midrule
1& {Local Query}&53.7&56.3&55.0 \\ 
2& {Global Query}&52.9&55.5&54.2 \\
3& {1+Cross-Frame}&56.2&59.1&57.6 \\
4& {1+Motion Collection}&58.1&61.5&59.8 \\
5& {3+4}&60.7&64.4&62.6 \\
6& {5+Gumble-Softmax}&61.4&65.4&63.4 \\
7& Ours &\textbf{63.6}&\textbf{68.0}   &\textbf{65.8}  \\  
\bottomrule[1pt]
\end{tabular}}
\end{center}
\caption{Ablation Study on Ref-Youtube-VOS dataset.}
\label{tab:ab}
\vspace{-0.4cm}
\end{table}

\subsection{Visualization}
\cc{Figure~\ref{fig:vz} visualizes several qualitative results. 
The referring expression in (a) describes the motion of a white duck, a pivotal feature that distinguishes it from similar ducks. Our proposed TempCD effectively captures the motion of the referent which is aligned with the language expression and enables accurate localization and segmentation of the specific duck. In contrast, Referformer fails to locate the referent object, primarily attributed to a semantic incongruity between the expression and the frame-level queries.
In the instance of (b), the deer's corresponding action appears in certain frames but lacks uniform presence. Our approach successfully captures its global semantics including motions and aligns it seamlessly with the referring expression, bridging the gap between local semantics and the referring expression via the Collection and Distribution mechanisms. Conversely, ReferFormer identifies the referent accurately in specific frames that correspond to the specified action, yet exhibits errors in others due to a noticeable lack of temporal context. All these observations collectively underscore the value of temporal interaction in fostering refined segmentation across frames.
Besides, (c) illustrates that our method can still precisely locate and segment referent objects in complex scenes. The second frame presents a scenario of occlusion, wherein the referent is obscured by another bicycle. Our approach detects the lack of correspondence between visible entities in a frame and the specified referent sequence, facilitated by cross-frame interactions between the visible objects and those delineated by the referent sequence. 
}

\section{Conclusion}
This paper proposes an end-to-end Temporal Query Collection and Distribution (TempCD) network for referring video object segmentation, which maintains object queries and the referent token and achieves alternating interaction between them via the proposed novel temporal collection-distribution mechanism. 

\noindent \textbf{Acknowledgment:} This work was supported by the National Natural Science Foundation of China (No.62206174), Shanghai Pujiang Program (No.21PJ1410900), Shanghai Frontiers Science Center of Human-centered Artificial Intelligence (ShangHAI), MoE Key Laboratory of Intelligent Perception and Human-Machine Collaboration (ShanghaiTech University), and Shanghai Engineering Research Center of Intelligent Vision and Imaging.

{\small
\bibliographystyle{ieee_fullname}
\bibliography{egbib}

\begin{thebibliography}{10}\itemsep=-1pt

\bibitem{botach2022end}
Adam Botach, Evgenii Zheltonozhskii, and Chaim Baskin.
\newblock End-to-end referring video object segmentation with multimodal transformers.
\newblock In {\em Proceedings of the IEEE/CVF Conference on Computer Vision and Pattern Recognition}, pages 4985--4995, 2022.

\bibitem{carion2020end}
Nicolas Carion, Francisco Massa, Gabriel Synnaeve, Nicolas Usunier, Alexander Kirillov, and Sergey Zagoruyko.
\newblock End-to-end object detection with transformers.
\newblock In {\em Computer Vision--ECCV 2020: 16th European Conference, Glasgow, UK, August 23--28, 2020, Proceedings, Part I 16}, pages 213--229. Springer, 2020.

\bibitem{chen2019see}
Ding-Jie Chen, Songhao Jia, Yi-Chen Lo, Hwann-Tzong Chen, and Tyng-Luh Liu.
\newblock See-through-text grouping for referring image segmentation.
\newblock In {\em Proceedings of the IEEE/CVF International Conference on Computer Vision}, pages 7454--7463, 2019.

\bibitem{chen2019referring}
Yi-Wen Chen, Yi-Hsuan Tsai, Tiantian Wang, Yen-Yu Lin, and Ming-Hsuan Yang.
\newblock Referring expression object segmentation with caption-aware consistency.
\newblock {\em arXiv preprint arXiv:1910.04748}, 2019.

\bibitem{cheng2021per}
Bowen Cheng, Alex Schwing, and Alexander Kirillov.
\newblock Per-pixel classification is not all you need for semantic segmentation.
\newblock {\em Advances in Neural Information Processing Systems}, 34:17864--17875, 2021.

\bibitem{ding2021vision}
Henghui Ding, Chang Liu, Suchen Wang, and Xudong Jiang.
\newblock Vision-language transformer and query generation for referring segmentation.
\newblock In {\em Proceedings of the IEEE/CVF International Conference on Computer Vision}, pages 16321--16330, 2021.

\bibitem{ding2022language}
Zihan Ding, Tianrui Hui, Junshi Huang, Xiaoming Wei, Jizhong Han, and Si Liu.
\newblock Language-bridged spatial-temporal interaction for referring video object segmentation.
\newblock In {\em Proceedings of the IEEE/CVF Conference on Computer Vision and Pattern Recognition}, pages 4964--4973, 2022.

\bibitem{feng2021encoder}
Guang Feng, Zhiwei Hu, Lihe Zhang, and Huchuan Lu.
\newblock Encoder fusion network with co-attention embedding for referring image segmentation.
\newblock In {\em Proceedings of the IEEE/CVF Conference on Computer Vision and Pattern Recognition}, pages 15506--15515, 2021.

\bibitem{gavrilyuk2018actor}
Kirill Gavrilyuk, Amir Ghodrati, Zhenyang Li, and Cees~GM Snoek.
\newblock Actor and action video segmentation from a sentence.
\newblock In {\em Proceedings of the IEEE Conference on Computer Vision and Pattern Recognition}, pages 5958--5966, 2018.

\bibitem{gori2005new}
Marco Gori, Gabriele Monfardini, and Franco Scarselli.
\newblock A new model for learning in graph domains.
\newblock In {\em Proceedings. 2005 IEEE International Joint Conference on Neural Networks, 2005.}, volume~2, pages 729--734. IEEE, 2005.

\bibitem{he2016deep}
Kaiming He, Xiangyu Zhang, Shaoqing Ren, and Jian Sun.
\newblock Deep residual learning for image recognition.
\newblock In {\em Proceedings of the IEEE conference on computer vision and pattern recognition}, pages 770--778, 2016.

\bibitem{hu2016segmentation}
Ronghang Hu, Marcus Rohrbach, and Trevor Darrell.
\newblock Segmentation from natural language expressions.
\newblock In {\em Computer Vision--ECCV 2016: 14th European Conference, Amsterdam, The Netherlands, October 11--14, 2016, Proceedings, Part I 14}, pages 108--124. Springer, 2016.

\bibitem{hu2020bi}
Zhiwei Hu, Guang Feng, Jiayu Sun, Lihe Zhang, and Huchuan Lu.
\newblock Bi-directional relationship inferring network for referring image segmentation.
\newblock In {\em Proceedings of the IEEE/CVF conference on computer vision and pattern recognition}, pages 4424--4433, 2020.

\bibitem{huang2020referring}
Shaofei Huang, Tianrui Hui, Si Liu, Guanbin Li, Yunchao Wei, Jizhong Han, Luoqi Liu, and Bo Li.
\newblock Referring image segmentation via cross-modal progressive comprehension.
\newblock In {\em Proceedings of the IEEE/CVF conference on computer vision and pattern recognition}, pages 10488--10497, 2020.

\bibitem{hui2021collaborative}
Tianrui Hui, Shaofei Huang, Si Liu, Zihan Ding, Guanbin Li, Wenguan Wang, Jizhong Han, and Fei Wang.
\newblock Collaborative spatial-temporal modeling for language-queried video actor segmentation.
\newblock In {\em Proceedings of the IEEE/CVF Conference on Computer Vision and Pattern Recognition}, pages 4187--4196, 2021.

\bibitem{hui2020linguistic}
Tianrui Hui, Si Liu, Shaofei Huang, Guanbin Li, Sansi Yu, Faxi Zhang, and Jizhong Han.
\newblock Linguistic structure guided context modeling for referring image segmentation.
\newblock In {\em Computer Vision--ECCV 2020: 16th European Conference, Glasgow, UK, August 23--28, 2020, Proceedings, Part X 16}, pages 59--75. Springer, 2020.

\bibitem{hwang2021video}
Sukjun Hwang, Miran Heo, Seoung~Wug Oh, and Seon~Joo Kim.
\newblock Video instance segmentation using inter-frame communication transformers.
\newblock {\em Advances in Neural Information Processing Systems}, 34:13352--13363, 2021.

\bibitem{jang2016categorical}
Eric Jang, Shixiang Gu, and Ben Poole.
\newblock Categorical reparameterization with gumbel-softmax.
\newblock {\em arXiv preprint arXiv:1611.01144}, 2016.

\bibitem{jhuang2013towards}
Hueihan Jhuang, Juergen Gall, Silvia Zuffi, Cordelia Schmid, and Michael~J Black.
\newblock Towards understanding action recognition.
\newblock In {\em Proceedings of the IEEE international conference on computer vision}, pages 3192--3199, 2013.

\bibitem{jiao2021two}
Yang Jiao, Zequn Jie, Weixin Luo, Jingjing Chen, Yu-Gang Jiang, Xiaolin Wei, and Lin Ma.
\newblock Two-stage visual cues enhancement network for referring image segmentation.
\newblock In {\em Proceedings of the 29th ACM International Conference on Multimedia}, pages 1331--1340, 2021.

\bibitem{jing2021locate}
Ya Jing, Tao Kong, Wei Wang, Liang Wang, Lei Li, and Tieniu Tan.
\newblock Locate then segment: A strong pipeline for referring image segmentation.
\newblock In {\em Proceedings of the IEEE/CVF Conference on Computer Vision and Pattern Recognition}, pages 9858--9867, 2021.

\bibitem{khoreva2019video}
Anna Khoreva, Anna Rohrbach, and Bernt Schiele.
\newblock Video object segmentation with language referring expressions.
\newblock In {\em Computer Vision--ACCV 2018: 14th Asian Conference on Computer Vision, Perth, Australia, December 2--6, 2018, Revised Selected Papers, Part IV 14}, pages 123--141. Springer, 2019.

\bibitem{kim2022restr}
Namyup Kim, Dongwon Kim, Cuiling Lan, Wenjun Zeng, and Suha Kwak.
\newblock Restr: Convolution-free referring image segmentation using transformers.
\newblock In {\em Proceedings of the IEEE/CVF Conference on Computer Vision and Pattern Recognition}, pages 18145--18154, 2022.

\bibitem{koner2022instanceformer}
Rajat Koner, Tanveer Hannan, Suprosanna Shit, Sahand Sharifzadeh, Matthias Schubert, Thomas Seidl, and Volker Tresp.
\newblock Instanceformer: An online video instance segmentation framework.
\newblock {\em arXiv preprint arXiv:2208.10547}, 2022.

\bibitem{li2020manigan}
Bowen Li, Xiaojuan Qi, Thomas Lukasiewicz, and Philip~HS Torr.
\newblock Manigan: Text-guided image manipulation.
\newblock In {\em Proceedings of the IEEE/CVF Conference on Computer Vision and Pattern Recognition}, pages 7880--7889, 2020.

\bibitem{li2018referring}
Ruiyu Li, Kaican Li, Yi-Chun Kuo, Michelle Shu, Xiaojuan Qi, Xiaoyong Shen, and Jiaya Jia.
\newblock Referring image segmentation via recurrent refinement networks.
\newblock In {\em Proceedings of the IEEE Conference on Computer Vision and Pattern Recognition}, pages 5745--5753, 2018.

\bibitem{liang2021clawcranenet}
Chen Liang, Yu Wu, Yawei Luo, and Yi Yang.
\newblock Clawcranenet: Leveraging object-level relation for text-based video segmentation.
\newblock {\em arXiv preprint arXiv:2103.10702}, 2021.

\bibitem{liang2021rethinking}
Chen Liang, Yu Wu, Tianfei Zhou, Wenguan Wang, Zongxin Yang, Yunchao Wei, and Yi Yang.
\newblock Rethinking cross-modal interaction from a top-down perspective for referring video object segmentation.
\newblock {\em arXiv preprint arXiv:2106.01061}, 2021.

\bibitem{lin2021structured}
Liang Lin, Pengxiang Yan, Xiaoqian Xu, Sibei Yang, Kun Zeng, and Guanbin Li.
\newblock Structured attention network for referring image segmentation.
\newblock {\em IEEE Transactions on Multimedia}, 24:1922--1932, 2021.

\bibitem{lin2017focal}
Tsung-Yi Lin, Priya Goyal, Ross Girshick, Kaiming He, and Piotr Doll{\'a}r.
\newblock Focal loss for dense object detection.
\newblock In {\em Proceedings of the IEEE international conference on computer vision}, pages 2980--2988, 2017.

\bibitem{liu2021cross}
Si Liu, Tianrui Hui, Shaofei Huang, Yunchao Wei, Bo Li, and Guanbin Li.
\newblock Cross-modal progressive comprehension for referring segmentation.
\newblock {\em IEEE Transactions on Pattern Analysis and Machine Intelligence}, 44(9):4761--4775, 2021.

\bibitem{liu2019roberta}
Yinhan Liu, Myle Ott, Naman Goyal, Jingfei Du, Mandar Joshi, Danqi Chen, Omer Levy, Mike Lewis, Luke Zettlemoyer, and Veselin Stoyanov.
\newblock Roberta: A robustly optimized bert pretraining approach.
\newblock {\em arXiv preprint arXiv:1907.11692}, 2019.

\bibitem{liu2021video}
Ze Liu, Jia Ning, Yue Cao, Yixuan Wei, Zheng Zhang, Stephen Lin, and Han Hu.
\newblock Video swin transformer.
\newblock {\em arXiv preprint arXiv:2106.13230}, 2021.

\bibitem{loshchilov2017adamw}
Ilya Loshchilov and Frank Hutter.
\newblock Decoupled weight decay regularization.
\newblock {\em arXiv preprint arXiv:1711.05101}, 2017.

\bibitem{luo2020multi}
Gen Luo, Yiyi Zhou, Xiaoshuai Sun, Liujuan Cao, Chenglin Wu, Cheng Deng, and Rongrong Ji.
\newblock Multi-task collaborative network for joint referring expression comprehension and segmentation.
\newblock In {\em Proceedings of the IEEE/CVF Conference on computer vision and pattern recognition}, pages 10034--10043, 2020.

\bibitem{maddison2016concrete}
Chris~J Maddison, Andriy Mnih, and Yee~Whye Teh.
\newblock The concrete distribution: A continuous relaxation of discrete random variables.
\newblock {\em arXiv preprint arXiv:1611.00712}, 2016.

\bibitem{milletari2016v}
Fausto Milletari, Nassir Navab, and Seyed-Ahmad Ahmadi.
\newblock V-net: Fully convolutional neural networks for volumetric medical image segmentation.
\newblock In {\em 2016 fourth international conference on 3D vision (3DV)}, pages 565--571. Ieee, 2016.

\bibitem{ning2020polar}
Ke Ning, Lingxi Xie, Fei Wu, and Qi Tian.
\newblock Polar relative positional encoding for video-language segmentation.
\newblock In {\em IJCAI}, volume~9, page~10, 2020.

\bibitem{qi2020reverie}
Yuankai Qi, Qi Wu, Peter Anderson, Xin Wang, William~Yang Wang, Chunhua Shen, and Anton van~den Hengel.
\newblock Reverie: Remote embodied visual referring expression in real indoor environments.
\newblock In {\em Proceedings of the IEEE/CVF Conference on Computer Vision and Pattern Recognition}, pages 9982--9991, 2020.

\bibitem{rezatofighi2019generalized}
Hamid Rezatofighi, Nathan Tsoi, JunYoung Gwak, Amir Sadeghian, Ian Reid, and Silvio Savarese.
\newblock Generalized intersection over union: A metric and a loss for bounding box regression.
\newblock In {\em Proceedings of the IEEE/CVF conference on computer vision and pattern recognition}, pages 658--666, 2019.

\bibitem{seo2020urvos}
Seonguk Seo, Joon-Young Lee, and Bohyung Han.
\newblock Urvos: Unified referring video object segmentation network with a large-scale benchmark.
\newblock In {\em Computer Vision--ECCV 2020: 16th European Conference, Glasgow, UK, August 23--28, 2020, Proceedings, Part XV 16}, pages 208--223. Springer, 2020.

\bibitem{chengshi}
Cheng Shi and Sibei Yang.
\newblock Spatial and visual perspective-taking via view rotation and relation reasoning for embodied reference understanding.
\newblock In {\em European Conference on Computer Vision}, pages 201--218. Springer, 2022.

\bibitem{shi2018key}
Hengcan Shi, Hongliang Li, Fanman Meng, and Qingbo Wu.
\newblock Key-word-aware network for referring expression image segmentation.
\newblock In {\em Proceedings of the European Conference on Computer Vision (ECCV)}, pages 38--54, 2018.

\bibitem{sreenu2019intelligent}
G Sreenu and Saleem Durai.
\newblock Intelligent video surveillance: a review through deep learning techniques for crowd analysis.
\newblock {\em Journal of Big Data}, 6(1):1--27, 2019.

\bibitem{cgformer}
Jiajin Tang, Ge Zheng, Cheng Shi, and Sibei Yang.
\newblock Contrastive grouping with transformer for referring image segmentation.
\newblock In {\em Proceedings of the IEEE/CVF Conference on Computer Vision and Pattern Recognition}, pages 23570--23580, 2023.

\bibitem{vaswani2017attention}
Ashish Vaswani, Noam Shazeer, Niki Parmar, Jakob Uszkoreit, Llion Jones, Aidan~N Gomez, {\L}ukasz Kaiser, and Illia Polosukhin.
\newblock Attention is all you need.
\newblock {\em Advances in neural information processing systems}, 30, 2017.

\bibitem{wang2020context}
Hao Wang, Cheng Deng, Fan Ma, and Yi Yang.
\newblock Context modulated dynamic networks for actor and action video segmentation with language queries.
\newblock In {\em Proceedings of the AAAI Conference on Artificial Intelligence}, volume~34, pages 12152--12159, 2020.

\bibitem{wang2019asymmetric}
Hao Wang, Cheng Deng, Junchi Yan, and Dacheng Tao.
\newblock Asymmetric cross-guided attention network for actor and action video segmentation from natural language query.
\newblock In {\em Proceedings of the IEEE/CVF International Conference on Computer Vision}, pages 3939--3948, 2019.

\bibitem{wang2021end}
Yuqing Wang, Zhaoliang Xu, Xinlong Wang, Chunhua Shen, Baoshan Cheng, Hao Shen, and Huaxia Xia.
\newblock End-to-end video instance segmentation with transformers.
\newblock In {\em Proceedings of the IEEE/CVF conference on computer vision and pattern recognition}, pages 8741--8750, 2021.

\bibitem{wang2022cris}
Zhaoqing Wang, Yu Lu, Qiang Li, Xunqiang Tao, Yandong Guo, Mingming Gong, and Tongliang Liu.
\newblock Cris: Clip-driven referring image segmentation.
\newblock In {\em Proceedings of the IEEE/CVF conference on computer vision and pattern recognition}, pages 11686--11695, 2022.

\bibitem{wu2020phrasecut}
Chenyun Wu, Zhe Lin, Scott Cohen, Trung Bui, and Subhransu Maji.
\newblock Phrasecut: Language-based image segmentation in the wild.
\newblock In {\em Proceedings of the IEEE/CVF Conference on Computer Vision and Pattern Recognition}, pages 10216--10225, 2020.

\bibitem{wu2022multi}
Dongming Wu, Xingping Dong, Ling Shao, and Jianbing Shen.
\newblock Multi-level representation learning with semantic alignment for referring video object segmentation.
\newblock In {\em Proceedings of the IEEE/CVF Conference on Computer Vision and Pattern Recognition}, pages 4996--5005, 2022.

\bibitem{wu2022language}
Jiannan Wu, Yi Jiang, Peize Sun, Zehuan Yuan, and Ping Luo.
\newblock Language as queries for referring video object segmentation.
\newblock In {\em Proceedings of the IEEE/CVF Conference on Computer Vision and Pattern Recognition}, pages 4974--4984, 2022.

\bibitem{wu2021seqformer}
Junfeng Wu, Yi Jiang, Wenqing Zhang, Xiang Bai, and Song Bai.
\newblock Seqformer: a frustratingly simple model for video instance segmentation.
\newblock {\em arXiv preprint arXiv:2112.08275}, 2021.

\bibitem{xu2015can}
Chenliang Xu, Shao-Hang Hsieh, Caiming Xiong, and Jason~J Corso.
\newblock Can humans fly? action understanding with multiple classes of actors.
\newblock In {\em Proceedings of the IEEE conference on computer vision and pattern recognition}, pages 2264--2273, 2015.

\bibitem{xu2021instance}
Zhujun Xu and Damien Vivet.
\newblock Instance sequence queries for video instance segmentation with transformers.
\newblock {\em Sensors}, 21(13):4507, 2021.

\bibitem{yang2019cross}
Sibei Yang, Guanbin Li, and Yizhou Yu.
\newblock Cross-modal relationship inference for grounding referring expressions.
\newblock In {\em Proceedings of the IEEE/CVF conference on computer vision and pattern recognition}, pages 4145--4154, 2019.

\bibitem{yang2019dynamic}
Sibei Yang, Guanbin Li, and Yizhou Yu.
\newblock Dynamic graph attention for referring expression comprehension.
\newblock In {\em Proceedings of the IEEE/CVF International Conference on Computer Vision}, pages 4644--4653, 2019.

\bibitem{yang2020propagating}
Sibei Yang, Guanbin Li, and Yizhou Yu.
\newblock Propagating over phrase relations for one-stage visual grounding.
\newblock In {\em Computer Vision--ECCV 2020: 16th European Conference, Glasgow, UK, August 23--28, 2020, Proceedings, Part XIX 16}, pages 589--605. Springer, 2020.

\bibitem{yang2020relationship}
Sibei Yang, Guanbin Li, and Yizhou Yu.
\newblock Relationship-embedded representation learning for grounding referring expressions.
\newblock {\em IEEE Transactions on Pattern Analysis and Machine Intelligence}, 43(8):2765--2779, 2020.

\bibitem{yang2022temporally}
Shusheng Yang, Xinggang Wang, Yu Li, Yuxin Fang, Jiemin Fang, Wenyu Liu, Xun Zhao, and Ying Shan.
\newblock Temporally efficient vision transformer for video instance segmentation.
\newblock In {\em Proceedings of the IEEE/CVF Conference on Computer Vision and Pattern Recognition}, pages 2885--2895, 2022.

\bibitem{yang2021bottom}
Sibei Yang, Meng Xia, Guanbin Li, Hong-Yu Zhou, and Yizhou Yu.
\newblock Bottom-up shift and reasoning for referring image segmentation.
\newblock In {\em Proceedings of the IEEE/CVF Conference on Computer Vision and Pattern Recognition}, pages 11266--11275, 2021.

\bibitem{yang2022lavt}
Zhao Yang, Jiaqi Wang, Yansong Tang, Kai Chen, Hengshuang Zhao, and Philip~HS Torr.
\newblock Lavt: Language-aware vision transformer for referring image segmentation.
\newblock In {\em Proceedings of the IEEE/CVF Conference on Computer Vision and Pattern Recognition}, pages 18155--18165, 2022.

\bibitem{ye2020dual}
Linwei Ye, Zhi Liu, and Yang Wang.
\newblock Dual convolutional lstm network for referring image segmentation.
\newblock {\em IEEE Transactions on Multimedia}, 22(12):3224--3235, 2020.

\bibitem{ye2019cross}
Linwei Ye, Mrigank Rochan, Zhi Liu, and Yang Wang.
\newblock Cross-modal self-attention network for referring image segmentation.
\newblock In {\em Proceedings of the IEEE/CVF conference on computer vision and pattern recognition}, pages 10502--10511, 2019.

\bibitem{yu2018mattnet}
Licheng Yu, Zhe Lin, Xiaohui Shen, Jimei Yang, Xin Lu, Mohit Bansal, and Tamara~L Berg.
\newblock Mattnet: Modular attention network for referring expression comprehension.
\newblock In {\em Proceedings of the IEEE conference on computer vision and pattern recognition}, pages 1307--1315, 2018.

\bibitem{yu2016modeling}
Licheng Yu, Patrick Poirson, Shan Yang, Alexander~C Berg, and Tamara~L Berg.
\newblock Modeling context in referring expressions.
\newblock In {\em Computer Vision--ECCV 2016: 14th European Conference, Amsterdam, The Netherlands, October 11-14, 2016, Proceedings, Part II 14}, pages 69--85. Springer, 2016.

\bibitem{zhao2022m4}
Wangbo Zhao, Kai Wang, Xiangxiang Chu, Fuzhao Xue, Xinchao Wang, and Yang You.
\newblock Modeling motion with multi-modal features for text-based video segmentation.
\newblock In {\em Proceedings of the IEEE/CVF Conference on Computer Vision and Pattern Recognition}, pages 11737--11746, 2022.

\bibitem{zhu2020deformable}
Xizhou Zhu, Weijie Su, Lewei Lu, Bin Li, Xiaogang Wang, and Jifeng Dai.
\newblock Deformable detr: Deformable transformers for end-to-end object detection.
\newblock {\em arXiv preprint arXiv:2010.04159}, 2020.

\end{thebibliography}
}

\end{document}